\documentclass{article}

\usepackage{arxiv}

\usepackage[utf8]{inputenc} 
\usepackage[T1]{fontenc}    
\usepackage{hyperref}       
\usepackage{url}            
\usepackage{booktabs}       
\usepackage{amsfonts}       
\usepackage{nicefrac}       
\usepackage{microtype}      
\usepackage{lipsum}
\usepackage{graphicx}
\graphicspath{ {./images/} }

\usepackage{amsmath}
\usepackage{amssymb}
\usepackage{amsthm}
\usepackage{multirow}
\usepackage{multicol}
\usepackage{subcaption}
\usepackage[ruled]{algorithm2e}
\DeclareMathOperator*{\topk}{argtopk\ }
\usepackage{tikz}

\newcommand{\pg}{\textsc{GAIM}}

\title{\pg: Attacking Graph Neural Networks via Adversarial Influence Maximization}

\author{
 Xiaodong Yang$^*$ \\
 Visa Research
   \And
 Xiaoting Li$^*$ \\
 Visa Research
  \And
 Huiyuan Chen \\
 Visa Research
 \And
 Yiwei Cai \\
 Visa Research
}

\begin{document}
\maketitle
\def\thefootnote{*}\footnotetext{These authors contributed equally to this work}
\begin{abstract}
Recent studies show that well-devised perturbations on graph structures or node features can mislead trained Graph Neural Network (GNN) models. However, these methods often overlook practical assumptions, over-rely on heuristics, or separate vital attack components. In response, we present \pg{}, an integrated adversarial attack method conducted on a node feature basis while considering the strict black-box setting. Specifically, we define an adversarial influence function to theoretically assess the adversarial impact of node perturbations, thereby reframing the GNN attack problem into the adversarial influence maximization problem. In our approach, we unify the selection of the target node and the construction of feature perturbations into a single optimization problem, ensuring a unique and consistent feature perturbation for each target node. We leverage a surrogate model to transform this problem into a solvable linear programming task, streamlining the optimization process. Moreover, we extend our method to accommodate label-oriented attacks, broadening its applicability. Thorough evaluations on five benchmark datasets across three popular models underscore the  effectiveness of our method in both untargeted and label-oriented targeted attacks. Through comprehensive analysis and ablation studies, we demonstrate the practical value and efficacy inherent to our design choices. 
\end{abstract}





\maketitle

\section{Introduction}
In recent years, Graph Neural Networks (GNNs) have emerged as increasingly powerful tools for graph understanding and mining \cite{abu2019mixhop,kipf2016semi,ying2018graph,liu2019higher,yu2024}. GNNs capitalize on the connectivity structure of graphs as a filter for aggregating neighborhood information, enabling them to extract high-level node features \cite{chen2018fastgcn}. As a consequence, GNNs have significantly advanced the state-of-the-art for various downstream tasks, including node classification and link prediction over graphs. 
However, despite their success, GNNs are not exempt from the inherent learning-security challenges found in standard machine learning models, such as a lack of adversarial robustness \cite{goodfellow2014explaining,szegedy2013intriguing}. A number of recent studies have illuminated the susceptibility of GNNs to adversarial attacks \cite{dai2018adversarial,wu2019adversarial,xu2019topology,zugner2018adversarial,zugner2019adversarial,li2023adversary,ma2022adversarial,ma2020towards,zang2020graph}. These attacks, armed with carefully crafted perturbations on graph structures and/or node features, can effectively deceive trained models.

Despite the considerable efforts, existing works have been subject to certain limitations. Such constraints include impractical assumptions about an attacker's knowledge of target models or their ability to manipulate data. Even the so-called restricted black-box attacks \cite{bojchevski2019adversarial,chang2020restricted,dai2018adversarial,chang2019general,ma2019attacking,ma2022adversarial}, where the attackers' access to a target model's parameters or outputs is limited, often permit access to model predictions or internal representations such as node embedding. Additionally, graph modifications should align with practical restrictions rather than assuming attackers can target any node at will. 
For example, nodes representing celebrities in social networks are often challenging to access or modify for significant attack impact. 
Moreover, most feature-level graph attacks exhibit limitations in the form of heuristic-based designs or the isolation of node selection and feature perturbation during the optimization process \cite{ma2019attacking,ma2022adversarial}. These steps are often treated independently, with one typically grounded in empirical methods. This disjointed approach hinders the pursuit of optimal solutions and diminishes the structural unity of the design.

In this study, we propose a new principled attack method on \textbf{G}raph Neural Networks via \textbf{A}dversarial \textbf{I}nfluence \textbf{M}aximization (\pg{}) to address the limitations at hand.
Our method is based on a greedy algorithm for this budgeted influence maximization problem, which iteratively selects the node that provides the maximum additional influence until the budget is exhausted.
We operate under a black-box framework where the attacker lacks access to fundamental model information such as parameters or predictions. In this context, we focus our research on node classification tasks and align with the course of adversarial attacks directed towards graph node features. Given a trained GNN and an input graph, the \pg{} attack aims to alter specific features across a limited number of identified nodes in order to undermine the GNN during the testing phase. 
In particular, we treat this attack as a problem of maximizing a node's adversarial influence. The primary objective is to pinpoint a small cluster of adversarially influential nodes and their corresponding feature perturbations, which can induce a substantial decline in the model's performance.


Departing from prior work, our methodology integrates the processes of target node selection and feature perturbation into a cohesive optimization strategy, ensuring desired attack outcomes. Specifically, we select an optimal subset of target nodes from our pool of candidates based on the computed adversarial influence scores. Concurrently, we calculate a unique feature perturbation for each chosen node, adhering strictly to our perturbation consistency protocol. This integrated approach provides a more comprehensive and effective strategy for launching adversarial attacks on GNNs. Nonetheless, solving the problem as formulated presents a challenge due to the inaccessibility of target models and the inherent nonlinearity of GNNs. In response, we employ a simplified surrogate model, SGC \cite{wu2019simplifying}, effectively transforming the problem into a linear programming (LP) issue, thereby drastically reducing computational complexity. At the same time, our experimental results highlight the strong transferability of the developed adversarial perturbations across models. This also aligns with insights unveiled from \cite{zugner2018adversarial,papernot2016transferability}, which demonstrate the famous concept of adversarial transferability between models.

Beyond the scope of this general untargeted attacks, we also seamlessly adapt our approach to two types of label-oriented targeted attacks, facilitating diverse attacking objectives. In our evaluation, we carry out extensive experimentation across five benchmark datasets over three common GNN models to assess the efficacy of \pg{} against the state-of-the-art techniques, we also evaluate the performance of our method under varying attack scenarios to validate its generalizability. 
Our findings consistently highlight the efficacy of our proposed approach within the landscape of GNN attacks. In summary, our contribution of this work can be summarized as follows:

\begin{itemize}
    \item We present a new and principled black-box adversarial attack on GNNs. It defines an adversarial influence function to capture the impact of node perturbations on the transmission of adversarial information among nodes. It transforms the problem of degrading the GNN classification performance to an adversarial influence maximization problem.
    \item We incorporate a surrogate model to convert the nonconvex adversarial influence maximization challenge into an LP problem, which enhances the efficiency of our optimization and enables its transferability to target models.
    \item We successfully adapt our method to facilitate two label-oriented attacks, enabling the achievement of a diverse range of attack objectives.
    \item Our method is compared with SOTA approaches, encompassing comprehensive analysis across five benchmark datasets and three widely-used GNN models. The evaluation results underscore its effectiveness.
\end{itemize}

\section{Related Work}
Graph Neural Networks (GNNs) are  known to be vulnerable to the quality of the input graphs
due to its recursive message passing schema~\cite{zugner2019adversarial,chen2022adversarial,jin2020adversarial,chen2021structured}.
The increasing incidence of adversarial attacks on GNNs is of significant concern as they pose a serious threat to the security and integrity of these networks.  Over recent years, numerous studies \cite{dai2018adversarial,wu2019adversarial,xu2019topology,zugner2018adversarial} have provided evidence of the vulnerability of GNNs to adversarial attacks.  For example, Nettack considers both test-time and training-time attacks on node classification
models~\cite{zugner2018adversarial}. These attacks, often achieved through minor perturbations to the graph structures or node features, can lead to severe misclassifications by the models. In light of these findings, the existing body of work can be classified based on several factors: the machine learning tasks, the intent of the attack, the phase in which the attack occurs, the nature of the attack, and the attacker's knowledge of the model during the attack process. Detailed analyses and comprehensive overviews of this literature are presented in numerous review papers \cite{jin2020adversarial,sun2022adversarial}.

The attack methodologies can be divided into two primary categories: (1) Graph Poisoning, which focuses on altering the original graph by introducing malicious modifications during the training phase to compromise the integrity and performance of the model \cite{bojchevski2019adversarial}, and (2) Graph Evasion, where attacks are conducted during the testing phase by subtly modifying the input data to evade detection or mislead the model without altering the underlying training graph \cite{zugner2019adversarial}. These distinct strategies highlight the different stages at which adversaries can target graph-based systems to achieve their malicious objectives. An alternative approach to classifying the existing work is to segregate them based on the attacker's knowledge of the target model.  White-box attacks \cite{xu2019topology,chen2019data} involve full knowledge of a system's internal workings, allowing for highly targeted and sophisticated attacks. In contrast, grey-box attacks \cite{zugner2018adversarial,sun2019node} are based on partial knowledge, and black-box attacks \cite{dai2018adversarial,chang2020restricted,ma2020towards,ma2022adversarial} rely solely on external interactions with no internal information, making them more challenging and reliant on trial and error. 

In this study,  we mainly focuses on evasion attacks and adheres to the black-box setting, which prevents any exploration of the model \cite{ma2022adversarial}. 
In particular, we formulate our attack as a Node Adversarial Influence Maximization problem and employ linear programming to improve the efficiency of the optimization process. Moreover, we integrate the selection of target nodes and their feature modification into a consolidated iterative optimization process. This offers enhanced assurances of attack competence and performance compared to the existing methods that treat these actions separately.  

\section{Preliminary}
\label{sec:prelim}

\subsection{Graph Neural Networks}
Many real-world data are naturally represented as graphs, such as social networks, gene interaction networks and citation relations among documents. Coping with those frequent applied tasks on graph data, GNNs provide powerful techniques for graph mining and understanding \cite{kipf2016semi,ying2018graph,liu2019higher}. Specifically, the GNNs use a well-designed message-passing scheme to perform nodes information aggregation via graph connectivity and extract high-level embeddings using message-passing schema. 

Without loss of generality, we denote an undirected graph as $G = (\mathcal{V}, \mathcal{E}, {X})$, where $\mathcal{V}$ ($|\mathcal{V}|=N$) is the set of nodes, $\mathcal{E}$ is the set of edges specifying node connections, and ${X} \in \mathbb{R}^{N \times D}$, where $D$ is the size of raw features. $A\in \{0,1\}^{N\times N}$ is the adjacency matrix.

A GNN model $f: \mathbb{R}^{N\times D} \rightarrow \mathbb{R}^{N\times K}$ is to map each node in graph to a label in $\mathcal{C}=\{c_1,c_2,\dots, c_K\}$. 
The computation of its $l$th layer can be formulated as,
\begin{equation}
\label{equ:layer}
    H^{(l)} = \text{ReLU}\Bigl(\hat{A} H^{(l-1)}W^{(l)}\Bigr),
\end{equation}
where $\hat{A} = \tilde{D}^{-\frac{1}{2}}\tilde{A}\tilde{D}^{-\frac{1}{2}}$ is a symmetrically normalized adjacency matrix and $\tilde{A}$ is the adjacency matrix $A$ added with self-loops and $\tilde{D}_{ii} = \sum_{j}\tilde{A}_{i,j}$. 
With $H^{0} = {X}$, the model with two layers can be formulated as, 
\begin{equation}
\label{equ:gcn}
    f_{\theta}(X,A) = \hat{A} \cdot \text{ReLU}\Bigl(\hat{A} X W^{(1)}\Bigr) W^{(2)},
\end{equation} 
where $\theta$ denotes the set of model parameters with $\theta = \{W^{(1)}, W^{(2)}\}$, $f_{\theta}(X,A)$ represents the output logits of the graph input. We denote $f_{\theta}^{j}(X,A) \in\mathbb{R}^K$ as the output logit of the node $j$ and $f_{\theta}^{j}(X,A)_c \in \mathbb{R}$ as the logit value corresponding to the label $c$. 

\subsection{GNN Adversarial Attacks}
The attack aims to identify representative nodes and their impactful features, such that conducting perturbation on them will result in the maximal performance degradation of models. The process includes the selection of nodes and features.

Similarly to works~\cite{ma2022adversarial,ma2020towards}, in the selection of nodes to perturb, the attack selects a small set of nodes $\mathcal{A}\subset \mathcal{V}$ under two constraints:
\begin{equation}
\label{equ:node_constr} 
    |\mathcal{A}| \leq B_n \And \forall v \in \mathcal{A},\   \text{deg}(v)\leq T
\end{equation}
where $B_n$ is the node budget, ${deg}$ denotes the degree of a node, and $T$ is predefined threshold. The purpose is to limit the target scope to ordinary nodes with small degrees and avoid the significant nodes being manipulated, which makes the attack practical. 

For the feature perturbation, the attacker picks out a small portion of impactful features of the selected nodes with a feature budget $B_f$, and then conducts perturbation $\epsilon_i$ on the node's feature $X_i\in \mathbb{R}^{1\times D}$, as follows: 
\begin{equation}
\label{equ:feature_constr}
     \mathcal{P}_i = \bigl\{\epsilon_i\ \big|\ lb\leq\ X_i + \epsilon_i \leq ub \And \|\epsilon_i\|_0 \leq B_f\bigr\},
\end{equation} where $ub$ and $lb$ denotes the upper bounds and lower bounds of features, respectively. It is noteworthy that $\hat{\epsilon}_{ij}\in\mathcal{P}_i$ will be used to represent the optimal feature perturbation derived by our method on the node $i$ for misclassifying its neighboring node $j$.

Overall, the goal of the attack is to find a subset of nodes where feature perturbation on them can lead to maximal misclassification of the other nodes. So the attack can be essentially formulated as an influence maximization problem:
\begin{equation}
\label{equ:max}
    \max_{\mathcal{A}} \big|\tau_G(\mathcal{A})\big|,\ \textbf{s.t.}\ \text{Equation}~(\ref{equ:node_constr}),~(\ref{equ:feature_constr}),
\end{equation}
where $|\tau_G|$ is the total number of nodes in the graph $G$ whose prediction is affected due to the feature perturbation on the selected nodes $\mathcal{A}$. 

\section{Methodology}
\label{sec:method}
\begin{figure*}[t]
	\centering
	\includegraphics[width=1.0\textwidth]{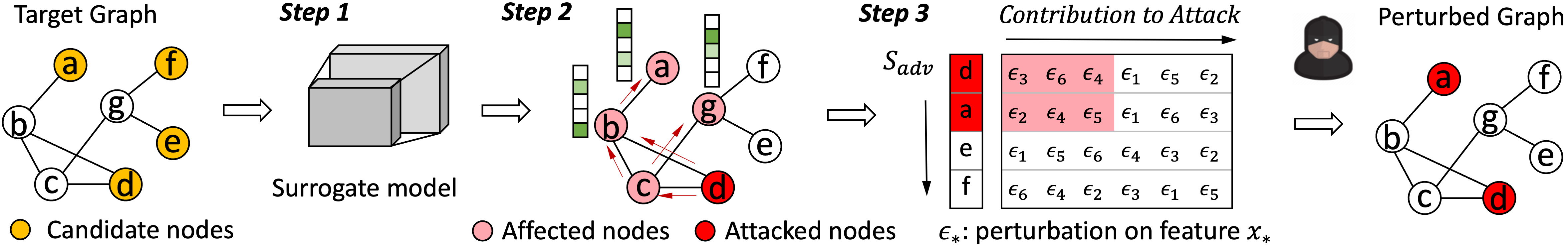}\\
	\caption{The framework of \pg{}, which mainly consists three steps: (1) Building a surrogate model; (2) Optimizing adversarial influence of candidate nodes; (3) Constructing feature perturbations and selecting target nodes}. \label{fig:framework}
\end{figure*}

This section introduces our attack method, \pg{}, which aims to identify the nodes with the highest adversarial influence by optimizing feature modifications. We achieve this by generating feature perturbations within the adversarial feature domain and calculating an adversarial influence score for each node. This score quantifies the node's ability to transmit misleading messages to other nodes. In our setting, a node's adversarial influence is defined as its capacity to induce prediction changes of its neighboring nodes under budgets. However, estimating this influence directly on a target GNN model can be challenging due to the inherent nonlinearity and the non-accessibility of the target model in restricted black-box attacks. As an alternative approach, we approximate the nodes' adversarial influence using a simplified surrogate model to mitigate complexity and non-accessibility. Leveraging the well-established concept of adversarial transferability between models, we utilize a specifically designed adversarial influence function to measure the impact of nodes on other nodes within the graph. By employing this framework, we can effectively identify the most influential nodes for attacks. 


We illustrate the framework of our method in Figure \ref{fig:framework}. The yellow circles in the figure represent candidate nodes whose degree is below a specified threshold. In Step 1, we train a surrogate model on the target graph. In Step 2, we compute the adversarial influences of each candidate node on the predictions of its neighboring nodes by modifying its features. To accomplish this, for each candidate node, we will identify a set of affected nodes denoted as $\mathcal{M}$, which are nodes whose predictions can be altered by applying perturbations to the candidate node within a predefined budget. 

The computation of these perturbations involves solving a series of optimization problems intersected with step 3. During this phase, we derive the optimal feature perturbation, represented as $\hat{\epsilon}_{ij} \in \mathcal{P}_i$ on the candidate node $i$ for its neighboring node $j$. The cumulation of these perturbations for a given candidate is denoted as a set $\Delta \mathcal{X}_i$. Subsequently, we rank the efficacy of each feature perturbation based on its contribution to the attack, retaining only the top-$B_f$ impactful ones. Following this, from all perturbations tailored for the candidate's neighbors, we identify the most prevalent perturbation patterns, forming the final perturbation for the candidate node. In selecting the target nodes, we apply this finalized feature perturbation, compute candidate influence scores, and select the top-$B_n$ nodes as our primary attack targets.



\subsection{Attacks $\&$ Adversarial Influence}
Harnessing the advantages of our highly integrated framework, our method enables the execution of three distinct attacks. The first one is the untargeted attacks~\cite{ma2022adversarial,ma2020towards}, which selects a group of target nodes to degrade the overall accuracy of a model. Complementing this approach, we also have developed two novel label-oriented attacks that have not been extensively studied before in the graph learning. Moreover, we emphasize the significance of perturbation consistency, an essential but often neglected aspect in untargeted attacks on graphs.

As described in Equation~(\ref{equ:node_constr}), the essence of these attacks is to identify a group of the most influential target nodes to maximize the attack effect.  As introduced above, the adversarial influence of one node is associated with its $\mathcal{M}$ that represent its neighboring nodes whose classification can be affected by perturbations on the node. 
Take a node $i$ for instance, given its neighboring nodes within hops $\mathcal{N}_i$, its $\mathcal{M}_i$ is formulated as:
\begin{equation}
\label{equ:M_i}
    \mathcal{M}_i = \bigl\{j\in {\mathcal{N}}_i \ \big|\  \exists \hat{\epsilon}_{ij}\in\mathcal{P}_i: \psi^{j}(X, A) \neq \psi^{j}(X +\hat{\epsilon}_{ij}, A) \bigr\}, 
\end{equation} where $\psi^{j}$ is the predicted label of the node $j$. The key part is the determination of the existence of the perturbation $\hat{\epsilon}_{ij}$. Our approach addresses this issue by transforming it into a series of optimization problems, where we maximize the worst-case margin between the logits corresponding to the initially predicted label and another label. 
More details will be discussed in the subsequent sections.

By solving these optimization problems, we can identify the optimal perturbation $\hat{\epsilon}_{ij}$ that maximizes the prediction loss of node $j$. If perturbation $\hat{\epsilon}_{ij}$ causes the misclassification of $j$, we will add the node $j$ to $\mathcal{M}_i$ and the corresponding perturbation $\hat{\epsilon}_{ij}$ will be included to $\Delta \mathcal{X}_i$ as
\begin{equation}
\label{equ:all_perturbs}
    \Delta\mathcal{X}_i = \{\hat{\epsilon}_{ij} \ |\ j\in \mathcal{M}_i\}.
\end{equation} Subsequently, we will aggregate all of these optimal perturbation to create a final perturbation for the node $i$.

As introduced, we propose three distinct attacks based on the aforementioned mechanism. One is an untargeted attack. The other two are label-orientated attacks. For these three attacks, the set $\mathcal{M}_i$ is adjusted by incorporating specific constraints to align with the attack objective.

\subsubsection{Untargeted Attack}
Untargeted attack methods such as~\cite{ma2022adversarial,ma2020towards} do not require the misclassified label $\psi^j(X+\hat{\epsilon}_{ij}, A)$ for each $j \in \mathcal{N}_i$ to be identical. That seems reasonable as the primary concern of adversarial influence lies in the capability to modify other nodes' predictions, without necessarily specifying the misclassified target label. However, a different scenario arises when crafting a node's feature perturbation by merging multiple impactful feature perturbations. As each of these features may yield to different misclassified labels, impulsively aggregating them could undermine the final perturbation's effectiveness in the attack. To ensure the attack's effectiveness, it is crucial to maintain what we refer to as \textbf{perturbation consistency}. In accordance with this principle, all perturbed features should lead to the same misclassified direction, implying that
$\psi^j(X+\hat{\epsilon}_{ij}, A)$ for each $j \in \mathcal{N}_i$ is the same.

Considering the perturbation consistency, the set $\mathcal{M}_{ic}$ with the label $c\in\mathcal{C}$ as the misclassified label is defined as
\begin{equation}
    \label{equ:untargetM}
    \mathcal{M}_{ic} = \mathcal{M}_i \cap \{j\in\mathcal{N}_i \ |\ \psi^j(X+\hat{\epsilon}_{ij}, A)=c\},
\end{equation} where $c$ is not necessarily the same for each candidate node $i\ \in \mathcal{V}$. The \textbf{adversarial influence score} of the node $i$ in our untargeted attack is computed by 
\begin{equation}
\label{equ:adversarial_influence}
    {S}_i =\max_{c\in {\mathcal{C}}}|\mathcal{M}_{ic}|.
\end{equation}

\subsubsection{Two Types of Label-oriented Targeted Attacks}
The adversarial influence score as described in Equation~(\ref{equ:adversarial_influence}) is employed in untargeted attacks with the objective of selecting target nodes to undermine the overall classification accuracy of models on graph nodes. In addition to this attack, we can expand this definition to encompass two types of label-oriented attacks. The first type involves modifying target nodes to diminish the classification accuracy for a specified label, denoted as $c_t$. In order to achieve this, the adversarial influence of a node must be assessed based on its capability to misclassify neighboring nodes that possess the target label. 
The node set $\mathcal{M}_i$ is accordingly modified to
\begin{equation}
    \text{Type I: } \mathcal{M}'_{ic_t} = \mathcal{M}_i\cap\bigl\{j\in {\mathcal{N}}_i\ \big|\ \psi^{j}(X, A) = c_t\bigr\},
\end{equation}
It is noteworthy that unlike $c$ in Equation~\ref{equ:untargetM}, the label $c_t$ remains fixed for all candidate nodes throughout the attack. Essentially, this attack aims to misclassify nodes who are already predicted as $\psi^{j}(X, A) = c_t$.

The other type of attack entails the selection of target nodes with the explicit objective of maximizing the misclassification rate towards a specific target label $c_t$. The goal is to induce the nodes to be erroneously classified as the target label. The $\mathcal{M}_i$ is updated to,
\begin{equation}
    \text{Type II: } \mathcal{M}''_{ic_t} = \mathcal{M}_i \cap \bigl\{j\in {\mathcal{N}}_i \ \big|\ \psi^{j}(X+\hat{\epsilon}_{ij}, A) = c_t\bigr\}.
\end{equation} 

To compute the adversarial influence score of node $i$ for these attacks, for each $j \in \mathcal{N}_i$, we need to verify whether there exists valid perturbation $\hat{\epsilon}_{ij} \in \mathcal{P}_i$ that affects its prediction, as formally described in the following problem.

Given a GNN model $f_{\theta}$, a graph $G$, a target node $i$ and one of its neighboring node $j$.  Let $\hat{c} \in \mathcal{C}$ be the node $j$'s predicted label with clean graph data. The worst-case margin between logit values of the labels $c$ ($c\in \mathcal{C}$ and $c\neq \hat{c}$) and $\hat{c}$ achievable under the perturbation domain $\mathcal{P}_i$ can be formulated by, 
\begin{equation}
\label{equ:opti}
\begin{aligned}
    p(c, \hat{c}):= \max_{\hat{\epsilon}_{ij} \in \mathcal{P}_i}\ f_{\theta}^{j}(X + \hat{\epsilon}_{ij}, A)_c - f_{\theta}^{j}(X+\hat{\epsilon}_{ij}, A)_{\hat{c}},
\end{aligned}
\end{equation}
where in the untargeted attack, $c$ can be any other labels, and in the two label-oriented attacks, $\hat{c}$ and $c$ should be the specified target label, respectively. If $\exists\hat{\epsilon}_{ij}\in \mathcal{P}_i \text{ such that } p(c, \hat{c})>0$, there exists perturbation in $\mathcal{P}_i$ that changes the prediction of node $j$ from the label $\hat{c}$ to the label $c$.

Addressing this optimization problem is significantly challenging. The primary difficulty lies in the nonlinearity of the objective function due to the presence of activation functions in the model. Notably, simplified graph convolutional networks (SGC) have demonstrated competitive performance despite being linear and lacking nonlinear activation functions. Furthermore, research has shown that adversarial examples exhibit high transferability between models~\cite{wu2019simplifying, demontis2019adversarial, zugner2018adversarial}. Based on these observations, we assume that the adversarial influence of nodes possesses considerable transferability across models, a hypothesis that we will validate through our experimental results. In light of this assumption, we propose estimating the adversarial influence of a node using a surrogate SGC model.

Our approach offers three notable advantages. Firstly, it transforms the nonlinear optimization problem into a linear programming (LP) problem, resulting in a significant reduction in computational complexity. Secondly, our attack operates as a black-box attack, eliminating the need for access to the model parameters. This characteristic enhances the practicality and applicability of our method in real-world scenarios where model parameters may not be readily available. Lastly, the adversarial influence estimated from surrogate models demonstrates a remarkable level of reusability. This transferability enables our attack to be employed against various models, including GCNs~\cite{kipf2016semi}, as well as models with different architectures such as JKNetMaxpool~\cite{xu2018representation} and GAT~\cite{velivckovic2017graph}.

\subsection{Adversarial Influence on A Surrogate Model} 
A SGC model is a linearized GCN model with the activation function $\text{ReLU}$ in Equation~(\ref{equ:layer}) being removed. Therefore, Equation~\ref{equ:layer} and \ref{equ:gcn} for the SGC model with $L$ layers are reformulated into:
\begin{equation*}
    H^{(l)} = \hat{A}H^{(l-1)}W^{(l)},\quad
    f_{\theta}(X,A) = \textbf{A} X\textbf{W},
\end{equation*} where $\textbf{A} = \hat{A}^{L}$ and $\textbf{W}=\prod_{l=1}^L W^{(l)}$
The output logits of a neighboring node $j$ with the perturbation $\hat{\epsilon}_{ij}$ on the node $i$ can be described as in Equation~(\ref{equ:linear_pert}), where $\alpha_{kj}$ is an entry of $\textbf{A}$ that represents the propagation weight between the nodes $k$ and $j$,  $\hat{\mathcal{N}}_j = \mathcal{N}_j \cup\{j\}$, and $[\textbf{W}]_c$ represents the column of $W$ corresponding to the label $c$.
\begin{equation}
\label{equ:linear_pert}
\begin{aligned}
    f_{\theta}^{j}(X+\hat{\epsilon}_{ij},A)_c & = \Biggl(\alpha_{ij}\hat{\epsilon}_{ij} + \sum_{k\in\hat{\mathcal{N}}_j}\alpha_{kj} X_k \Biggr) [\textbf{W}]_c
\end{aligned}
\end{equation} 

Therefore, by substituting the $f_{\theta}^{j}(X+\hat{\epsilon}_{ij},A)$ in Equation~(\ref{equ:opti}) with the expression in Equation~(\ref{equ:linear_pert}), we effectively convert the original nonlinear optimization problem that aims to change the predicted label of the node $j$ from $\hat{c}$ to $c$ 
into an LP problem:
\begin{equation}
\label{equ:opti_integer}
\begin{aligned}
    p(c,\hat{c})&:=\max_{\hat{\epsilon}_{ij} \in \mathcal{P}_i}\ \Biggl(\alpha_{ij}\hat{\epsilon}_{ij} + \sum_{k\in\hat{\mathcal{N}}_j}\alpha_{kj} X_k \Biggr)\bigl([\textbf{W}]_c - [\textbf{W}]_{\hat{c}}\bigr)
\end{aligned}
\end{equation}

\subsubsection{Solving of The LP Problem} 
\label{sssec:solve_lp}
In Equation~(\ref{equ:feature_constr}), the feasible region $\mathcal{P}_i$ is a conjunction of $lb \leq X_i + \epsilon_i \leq ub$ and $\|\epsilon_i\|_0 \leq B_f$. The domain $lb \leq X_i + \epsilon_i \leq ub$ is essentially a box constraint. Let $x_{i,d}$ denote the $d$th entry of $X_i$, then, we have the range of $\hat{\epsilon}_{ij,d}$ to be,
\begin{equation*}
    lb_d-x_{i,d} \le \hat{\epsilon}_{ij,d} \le ub_d - x_{i,d}.
\end{equation*} and the domain $\|\epsilon_i\|_0 \leq B_f$ specifies the maximal number of features to perturb. 
We first expand the objective function into 
\begin{equation*}
\begin{aligned}
    p(c,\hat{c})&:=\max_{\hat{\epsilon}_{ij} \in \mathcal{P}_i}\  \sum_{d=1}^{D}\omega_d\hat{\epsilon}_{ij,d}  + \sum_{k\in\hat{\mathcal{N}}_j}\alpha_{kj} \Biggl( \sum_{d=1}^{D} x_{k,d}(w_{dc} -w_{d\hat{c}}) \Biggr),
\end{aligned}
\end{equation*}
where $\omega_d = \alpha_{ij}(w_{dc} -w_{d\hat{c}})$ and $w_{dc}$ is the entry of $\textbf{W}$. We can notice that  $\omega_d$ is the coefficient, $\hat{\epsilon}_{ij,d}$ is the decision variable, and the item after the addition sign is a constant. When $\omega_d$ is positive and negative, we have
\begin{align*}
    \omega_d (lb_d-x_{i,d}) \leq \omega_d\hat{\epsilon}_{ij,d}\leq \omega_d(ub_d-x_{i,d}),\\ \omega_d (ub_d-x_{i,d}) \leq \omega_d\hat{\epsilon}_{ij,d}\leq \omega_d(lb_d-x_{i,d}),
\end{align*} respectively. Therefore, the optimal solution of $p(c,\hat{c})$ over the constraint $lb \leq X_i + \epsilon_i \leq ub$ is that
\begin{equation}
    p(c,\hat{c}) = \sum_{d=1}^{D}\omega_d\hat{\epsilon}_{ij,d} + \sum_{k\in\hat{\mathcal{N}}_j}\alpha_{kj} \Biggl( \sum_{d=1}^{D} x_{k,d}(w_{dc} -w_{d\hat{c}}) \Biggr),
\end{equation} where
\begin{equation}
\label{equ:pert}
    \hat{\epsilon}_{ij,d} = \begin{cases}
       ub_d-x_{i,d} & \text{ if } \omega_d >0 \\
       lb_d-x_{i,d} & \text{ otherwise }
    \end{cases}
\end{equation}

For the constraint $\|\epsilon_i\|_0 \leq B_f$ that specifies the maximal number of features to perturb, 
the top $B_f$ features whose $\omega_d \hat{\epsilon}_{ij,d} $ positively contributes most to the optimal solution are selected to approach the optimal solution of $p(c,\hat{c})$ over $\mathcal{P}_i$. We treat these feature perturbations as important ones. They are,
\begin{equation}
    \mathcal{F}_i = \Biggl(\topk_{k=B_f} \bigcup_{d=1}^{D} \omega_d  \hat{\epsilon}_{ij,d} \Biggr) \cap \bigl\{\ d\ \big|\ \omega_d \hat{\epsilon}_{ij,d} >0\bigr\}. 
\end{equation}
Subsequently, we can obtain the perturbation for each selected features in $\mathcal{F}_i$ in terms of Equation~(\ref{equ:pert}). It repeatedly computes all the optimal perturbations $\hat{\epsilon}_{ij}$ for the neighboring nodes of node $i$ under budget and forms the perturbation set $\Delta\mathcal{X}_i$, which we will use for final perturbation construction.

It should be underscored that, apart from dealing with continuous features, our methodology is also flexible enough to accommodate binary features within the graph node. This capability stems from the fact that the perturbed feature aligns with either the upper or lower bound of its value range, as defined by Equation~(\ref{equ:pert}). The initial step involves relaxing the binary discrete domain, $\{0,1\}$, into a continuous domain, $[0,1]$. Once our method is applied, the perturbed feature values gravitate towards either the lower bound $0$ or the upper bound $1$, thus preserving its binary character. The same applies to the bounded discrete range.

\subsubsection{Construction of Final Perturbation}
The aforementioned computation of the optimal perturbation on the node $i$ for each of its neighboring nodes yield a set of perturbation denoted as $\Delta \mathcal{X}_i$ in Equation~(\ref{equ:all_perturbs}). How to craft the final perturbation for $i$ that maximizes its overall adversarial influence is critical. Empirically, we observe that in our attack the selected features and their perturbations of each $\hat{\epsilon}_{ij}$ in $\Delta \mathcal{X}_i$ share a lot of similarity. Therefore, we utilize the most common features and their perturbations in $\Delta \mathcal{X}_i$ to form the final perturbation. This crafting process is efficient and its effectiveness is also validated in our experiments.

\subsubsection{The Black-box GNN Attack Procedure}

In a nutshell, we operate \pg{} for maximizing adversarial influence on nodes by combining target node selection with optimized feature perturbations in a black-box setup. Initially, candidate nodes are identified by filtering out those with degrees exceeding the threshold (Equation~(\ref{equ:node_constr})). For each candidate, we compute influence scores and feature perturbations using the method from Section~\ref{sec:method}. Notably, a neighboring node's use in one candidate's influence score computation precludes its reuse for others, maximizing overall impact as in Equation~(\ref{equ:max}). Subsequently, we select the most impactful candidate nodes with perturbations to deploy the modified graph for testing. 

\begin{table}[h]
    \centering
    \caption{Dataset statistics} \label{tab:datasets}
    \tabcolsep=3.5pt
    \begin{tabular}{lcccccc}
        \toprule
         Dataset & Cora & Citeseer & Pubmed & Flickr & Reddit  \\
        \midrule
        \# Nodes & 2,485 & 3,327 & 19,717 & 89,250 & 232,965\\
        \# Edges & 10,138 & 4,732 & 88,648 & 899,756 & 11,606,919\\
        \# Features & 1,433 & 3,703 & 500 & 500 & 602\\
        \# Classes & 7 & 6 & 3 & 7 & 41 \\
        Feature type & Binary & Binary & Binary & Discrete & Continuous\\
        \bottomrule
\end{tabular} 
\end{table}

\section{Experimental Results and Analysis}

This section presents a comprehensive evaluation of our proposed attack strategy (\pg{}) against several typical GNN models, and compare its effectiveness with the baseline methods. Additionally, we conduct the parameter analysis to study their impacts on the attack performance, and ablation study to investigate the importance of different components in our design.

\begin{table*}[h]
\centering
\caption{Attack performances in terms of test accuracy (\%) on small datasets. Lower values indicate better performance. The error bar $\pm$ denotes the standard deviation of the results in each setup.}
\label{tab:small_table}
{
\renewcommand{\arraystretch}{1.1}
\tabcolsep=3pt
\begin{tabular}{lccccccccc}
\toprule
\multicolumn{1}{l|}{\multirow{2}{*}{Method}} & \multicolumn{3}{c|}{Cora}                                                       & \multicolumn{3}{c|}{Citeseer}                                             & \multicolumn{3}{c}{Pubmed}                                  \\
\multicolumn{1}{l|}{}                        & JKNetMax          & GCN                  & \multicolumn{1}{c|}{GAT}             & JKNetMax          & GCN               & \multicolumn{1}{c|}{GAT}          & JKNetMax          & GCN               & \multicolumn{1}{c}{GAT}              \\ \hline
\multicolumn{1}{l|}{None}                    & 88.0$\pm$1.4      & 86.3$\pm$1.9         & \multicolumn{1}{c|}{85.0$\pm$1.8}    & 75.9$\pm$1.4      & 74.9$\pm$1.5      & \multicolumn{1}{c|}{74.6$\pm$1.5} & 86.1$\pm$0.6      & 85.0$\pm$0.4      & \multicolumn{1}{c}{85.1$\pm$0.5}   \\ 
\multicolumn{1}{l|}{Random}                  & 79.5$\pm$2.5      & 82.6$\pm$2.0         & \multicolumn{1}{c|}{71.6$\pm$5.6}    & 70.5$\pm$2.0      & 70.9$\pm$1.6      & \multicolumn{1}{c|}{68.8$\pm$2.1} & 79.1$\pm$0.8      & 81.3$\pm$0.6      & \multicolumn{1}{c}{66.1$\pm$2.4} \\
\multicolumn{1}{l|}{Degree}                  & 72.3$\pm$2.3      & 80.0$\pm$2.3         & \multicolumn{1}{c|}{64.1$\pm$7.3}    & 61.6$\pm$2.3      & 65.8$\pm$1.9      & \multicolumn{1}{c|}{61.6$\pm$2.0} & 66.1$\pm$1.2      & 74.3$\pm$1.3      & \multicolumn{1}{c}{53.7$\pm$2.0} \\
\multicolumn{1}{l|}{Pagerank}                & 79.1$\pm$2.1      & 81.9$\pm$1.7         & \multicolumn{1}{c|}{70.9$\pm$4.8}    & 65.8$\pm$1.7      & 68.6$\pm$1.7      & \multicolumn{1}{c|}{65.1$\pm$1.8} & {70.5$\pm$1.2}    & {76.6$\pm$1.2}    & \multicolumn{1}{c}{58.7$\pm$1.6} \\
\multicolumn{1}{l|}{Betweenness}             & 80.4$\pm$2.0      & 81.9$\pm$1.7         & \multicolumn{1}{c|}{74.0$\pm$3.9}    & 69.3$\pm$1.7      & 70.2$\pm$1.6      & \multicolumn{1}{c|}{68.5$\pm$2.0} & 74.6$\pm$1.0      & 79.2$\pm$0.9      & \multicolumn{1}{c}{59.3$\pm$2.0} \\

\multicolumn{1}{l|}{RWCS}                    & 79.9$\pm$2.0      & {80.7$\pm$2.0}       & \multicolumn{1}{c|}{66.4$\pm$3.1}    & 69.2$\pm$1.6      & 69.9$\pm$1.5      & \multicolumn{1}{c|}{69.0$\pm$1.6} & {82.3$\pm$0.9}    & {80.7$\pm$0.4}    & \multicolumn{1}{c}{78.8$\pm$1.3}\\
\multicolumn{1}{l|}{GC-RWCS}                 & 80.0$\pm$2.7      & 83.1$\pm$2.0         & \multicolumn{1}{c|}{54.4$\pm$3.0}    & 59.6$\pm$2.3      & 65.8$\pm$2.2      & \multicolumn{1}{c|}{60.1$\pm$2.5} & 83.8$\pm$1.0      & 82.1$\pm$0.6      & \multicolumn{1}{c}{78.9$\pm$2.2}  \\ 
\multicolumn{1}{l|}{InfMax-Unif}             & 79.9$\pm$2.8      & 83.5$\pm$2.1         & \multicolumn{1}{c|}{\textbf{52.7$\pm$2.8}}  & 59.5$\pm$2.3      & 65.1$\pm$2.1      & \multicolumn{1}{c|}{60.2$\pm$2.4}  & 83.9$\pm$1.0     & 81.8$\pm$0.5      & \multicolumn{1}{c}{78.5$\pm$2.1}  \\ 
\multicolumn{1}{l|}{InfMax-Norm}             & 80.1$\pm$2.8      & 83.6$\pm$1.9         & \multicolumn{1}{c|}{{54.3$\pm$2.7}}  & 59.2$\pm$2.5      & 64.9$\pm$2.4      & \multicolumn{1}{c|}{60.0$\pm$2.3}  & 84.0$\pm$1.0     & 81.8$\pm$0.6      & \multicolumn{1}{c}{78.1$\pm$2.4}\\
\multicolumn{1}{l|}{{\pg{} (Ours.)}}                 & \textbf{64.6$\pm$1.8}      & \textbf{67.6$\pm$2.1}       & \multicolumn{1}{c|}{{61.2$\pm$3.2}}  & \textbf{51.1$\pm$1.6}      & \textbf{57.1$\pm$2.1}      & \multicolumn{1}{c|}{\textbf{55.8$\pm$1.9}} & \textbf{52.4$\pm$1.2}    & \textbf{61.1$\pm$0.9}    & \multicolumn{1}{c}{\textbf{47.8$\pm$1.4}} \\ 
\bottomrule
\end{tabular}
}
\end{table*}

\begin{table*}[h]
\centering
\caption{Attack performances in terms of test accuracy (\%) on large datasets. Related works RWCS, GC-RWCS, InfMax-Unif and InfMax-Norm are not implemented due to their scalability issues.}
\label{tab:large_table}
{
\renewcommand{\arraystretch}{1.1}
\tabcolsep=8.0pt
\begin{tabular}{lcccccc}
\toprule
\multicolumn{1}{l|}{\multirow{2}{*}{Method}}    & \multicolumn{3}{c|}{Flickr}    & \multicolumn{3}{c}{Reddit}                         \\
\multicolumn{1}{l|}{}        & JKNetMax          & GCN          & \multicolumn{1}{c|}{GAT}            & JKNetMax      & GCN               & GAT           \\ \hline
\multicolumn{1}{l|}{None}                    & 51.0$\pm$0.6      & 51.9$\pm$0.5 & \multicolumn{1}{c|}{52.3$\pm$0.7}   &{94.5$\pm$0.1} & 94.1$\pm$0.1      & 93.4$\pm$0.2    \\ 
\multicolumn{1}{l|}{Random}                  & 47.6$\pm$1.0      & 48.0$\pm$1.3 & \multicolumn{1}{c|}{44.2$\pm$7.4}   &{94.2$\pm$0.2} & 93.8$\pm$0.5      & 44.5$\pm$19.6  \\ 
\multicolumn{1}{l|}{Degree}                  & 46.0$\pm$1.2      & 47.0$\pm$1.5 & \multicolumn{1}{c|}{43.0$\pm$8.3}   &{94.1$\pm$0.2} & 93.7$\pm$0.6      & 33.1$\pm$10.3   \\ 
\multicolumn{1}{l|}{Pagerank}                & 47.8$\pm$1.6      & 47.3$\pm$1.6 & \multicolumn{1}{c|}{47.0$\pm$4.8}   &{92.2$\pm$0.6} & 92.4$\pm$0.5      & 47.7$\pm$14.7   \\ 
\multicolumn{1}{l|}{Betweenness}             & 47.2$\pm$2.1      & 46.1$\pm$2.0 & \multicolumn{1}{c|}{46.0$\pm$5.5}   &{92.8$\pm$0.4} & 92.8$\pm$0.5      & 44.8$\pm$14.5  \\ 
\multicolumn{1}{l|}{{\pg{} (Ours.)}}         & \textbf{44.0$\pm$1.3}      & \textbf{44.5$\pm$1.5} & \multicolumn{1}{c|}{\textbf{40.5$\pm$4.2}}   &\textbf{75.0$\pm$0.7} & \textbf{73.1$\pm$1.2}      & \textbf{16.4$\pm$13.5} \\ 
\bottomrule
\end{tabular}
}
\end{table*}

\subsection{Experimental Setup}\label{sec:setup}
\textbf{Datasets and GNN models.} In this study, we use node classification tasks to assess the attack capability of \pg{} on five benchmark datasets: Cora, Citeseer, Pubmed \cite{sen2008collective}, 
one online image network Flickr \cite{zeng2019graphsaint} and Reddit \cite{hamilton2017inductive}. A summary of the basic properties of these datasets is provided in Table~\ref{tab:datasets}.
In all of our experiments, we randomly split each dataset at the ratio of $3:1:1$ for training, validation, and testing. For each attack setting, we run 20 trials and gather the average as our results. 
We evaluate our attack strategy on several commonly-used GNN models, including: (1) GCN \cite{kipf2016semi}, (2) JK-NetMaxpool \cite{xu2018representation}, and (3) GAT \cite{velivckovic2017graph}. We set the number of layers to 2 for all models and set the number of heads to 8 for GAT. The complete experimental result is available at our supplementary material.

\textbf{Baselines}
We compare our strategy with two distinct groups of methods for fair comparison. The first group contains the popular heuristic-based metrics commonly used to measure the informativeness of nodes in few-shot learning or Active Learning problems \cite{settles2009active,aggarwal2014active}. 
These methods are typically employed as selection criteria to identify representative nodes in graph learning, based on the intuition that they have a greater impact on other nodes in the message-passing scheme.

In this study, we use three well-known baselines from this group, namely \textbf{Degree}, \textbf{PageRank}, and \textbf{Betweenness}. Additionally, we include \textbf{Random} selection as a trivial baseline. In the second group, we adopt the methods in the related works \cite{ma2020towards,ma2022adversarial} to compare with. In these methods, \textbf{RWCS} and \textbf{GC-RWCS} \cite{ma2020towards} derive adversarial attacks by approximately maximizing the cross-entropy classification loss using heuristics, while \textbf{InfMax-Unif} and \textbf{InfMax-Norm} \cite{ma2022adversarial} model the problem of maximizing model misclassification rate as an influence maximization problem on a variant of linear threshold model. For this group of baselines, we replicated the parameter setup as described in the original works.

Notably, all the baseline strategies solely focus on target node selections and neglect the perturbation construction on node features. This emphasizes the uniqueness of our design. In constructing the perturbation vector in these baseline methods, we adopt the approach outlined in the paper \cite{ma2022adversarial}. They train $20$ GCN models as proxy models and compute the average gradients of the classification loss with respect to the node features. Then, the features with the top gradients are selected to form a global perturbation applied to all selected nodes. This perturbation is set with a modification length using the sign of the gradients. 

In contrast, our approach takes a more unified and data-agnostic perspective. For each target node, we create the perturbation from multiple optimal feature perturbations to maximize its adversarial influence on neighboring nodes. This customized feature perturbation approach sets our method apart from the baseline strategies and enhances its effectiveness in target node selections while considering the specific characteristics of each individual node and its influence on the graph structure.


\subsection{Method Evaluation}
\label{sec:eval}

In this part, we evaluate the performance of our method under different settings. In practical adversarial scenarios, ensuring the imperceptibility of the attack is a crucial design requirement \cite{szegedy2013intriguing}. To achieve feasibility and practicality, we impose three essential constraints on our attack strategy. (1) We set the maximum number of target nodes to be only $1\%$ of the total graph size for the datasets Cora, Citeseer and Pubmed, and $0.1\%$ for the larger datasets Flickr and Reddit. 
(2)We perform perturbations exclusively on nodes with low degrees.
, as they are typically more accessible in real-world scenarios. In contrast, high-degree nodes are unlikely to have their properties altered, rendering them unrealistic targets.
We restrict our selection of candidate nodes to be the set after removing top $10\%$ and $30\%$ of nodes with the highest degree; (3) The feature modification rate of the target nodes is set as $2\%$ for Cora, Citeseer and Pubmed, and $5\%$ for Flickr and Reddit. These controlled rates ensure that the alterations to the nodes' features remain subtle and inconspicuous while still exerting significant influence. 
The features of Flickr and Reddit are actually embedding vectors which don't have specific ranges. Therefore, we set their upper bound and lower bound with the global maximum and minimum value over all features in the data, respectively.

\subsubsection{Comparison results under untargeted attack settings.}
In this section, we evaluate our proposed attack method against baseline techniques in a general untargeted attack setting. 
The objective was to validate the effectiveness of our approach, and to this end, we collected attack results for three models across five datasets. The evaluation results are summarized in Table~\ref{tab:small_table} and~\ref{tab:large_table}, which presents the results with removing top $10\%$ nodes with highest degree. For the results pertaining to the removal of $30\%$ of nodes are detailed in our supplementary material. In the table, the label \emph{None} denotes the regular GNN with no attacks applied. As the results illustrate, our attack strategy significantly diminishes the classification accuracy of diverse GNN models across all the experimented datasets. We managed to induce an average reduction in accuracy of $21.0\%$ on the Cora dataset, $21.7\%$ on Citeseer, $13.0\%$ on Pubmed, $8.7\%$ on Flickr, and a striking $39.2\%$ on Reddit. We think the substantial performance degradation on Reddit could be attributed to the wide range of feature boundaries, allowing for large perturbations.

Our method consistently outperformed most of the baseline methods for all attacking settings. Particularly, our approach exhibited superior performance compared to others when applied to the JKNetMax model and GCN model. 
It is important to note that baseline methods (\textit{RWCS, GC-RWCS, InfMax-Unif, InfMax-Norm}) face challenges when scaling to large graph datasets and are restricted in their applicability to black-box scenarios. In contrast, our proposed method is not encumbered by such constraints and can be proficiently deployed across various scenarios. Remarkably, 
As shown in Table~\ref{tab:large_table}, our method outperforms the best-performing baseline across the three models with two large datasets. This solid empirical performance demonstrates the efficacy of our attack methodology and affirms its potential to hinder the reliability of GNN models across a spectrum of datasets.

\subsubsection{Evaluation results of Label-oriented attacks.} 
\begin{figure*}[t!]
    \centering
    \includegraphics[width=0.98\textwidth]
    {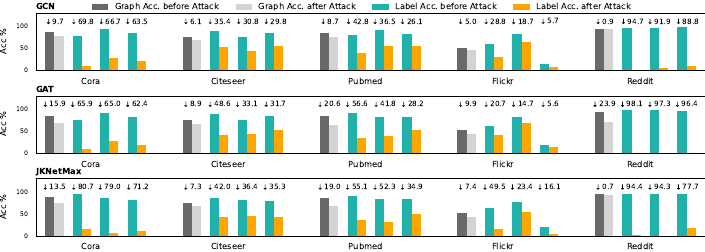}
    \caption{Type-I Label-orientated Attack. It aims to degrade the accuracy of target labels. Top 3 target labels are displayed.}
    \label{fig:typeI}
\end{figure*}

In this section, we present the evaluation results for two types of label-oriented attack settings. We conduct the experiments on JKNetMax, GCN, and GAT models across all datasets. The first type involves degrading the model's performance specifically on the specified label. {Hyperparameters keep the same as before}. During testing, we measure both the accuracy of the attacking label and the overall performance of the model across all labels, recording the outcomes for analysis. We display the top three labels which are mostly impacted by attack in Figure~\ref{fig:typeI} Type-I attack. The results show that our method significantly reduces the classification accuracy of the targeted labels.
Remarkably, our technique manages to degrade the targeted classification of these three models by an average of $69.4\%$ for Cora, $35.9\%$ for Citeseer, $41.6\%$ for Pubmed, $20.4\%$ for Flickr and 
and $92.6\%$ for Reddit. The substantial drops in Reddit is mainly due to the large range of feature values which enables strong attacks. Overall, these results validate our approach's capability of strategically compromising the model's predictions for specific labels, highlighting its utility for label-oriented attacks

\begin{figure*}
    \centering
    \includegraphics[width=0.98\textwidth]{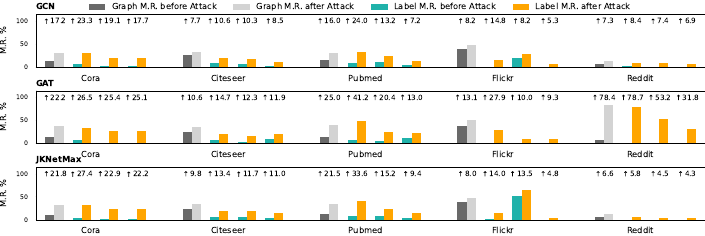}
    \caption{Type-II Label-orientated Attack. It aims to increase the ratio of nodes being misclassified into target labels. M.R. stands for misclassification rate. Top 3 target labels are displayed.}
    \label{fig:typeII}
\end{figure*}

The second type of label-oriented attack seeks to induce the model into misclassifying nodes as the targeted label. We compute the misclassification rate for each label during testing, representing the proportion of nodes successfully misclassified as the target label. In Figure~\ref{fig:typeII} Type-II attack, we display the three labels with the highest misclassification rate, along with the overall misclassification rate over all labels. The results indicate that our method achieves obvious misclassification enhancement for the targeted labels. This type of attack can also result in high misclassification rates for all other labels. 
However, the results for targeted labels are slightly less significant compared to Type-I attack, due to the inherent difficulty of misclassifying a node to a specified label, as opposed to misclassifying a node with a specific label to any other label. Nevertheless, our findings underscore the effectiveness of our attack strategy in manipulating the model to misclassify nodes, which further highlights its utility for label-oriented attacks designed to change model's predictions in a controlled manner. 

\subsection{Parameter Analysis}

In this section, we investigate the impact of the two hyperparameters, namely the node budget $B_n$ and the feature budget $B_f$, on the performance of our method. Notely, our approach does not require additional hyperparameters. The node budget represents the percentage of target nodes related to the graph size, while the feature budget is the percentage of the maximum allowed modified features in each node.

To ensure a comprehensive analysis and fair comparison, we divide the investigation into two parts based on the sizes of different datasets. When we study the impact of different node budget, we vary the node budget $B_n$ over the values $\{1\%, 2\%, 3\%, 4\%, 5\%\}$ for smaller graph datasets (Cora, Citeseer, Pubmed), with the feature budget $B_f$ remaining constant at $2\%$. For the larger graph datasets (Flickr and Reddit), $B_n$ is varied across $\{0.1\%, 0.2\%, 0.3\%, 0.4\%, 0.5\%\}$, with the $B_f$ fixed at $5\%$. As we analyze the influence of different feature budgets $B_f$, we adjust $B_f$ to span $\{1\%, 2\%, 3\%, 4\%, 5\%\}$ for smaller graphs, and $\{5\%, 10\%, 15\%, 20\%, 25\%\}$ for the two larger graphs. The node budgets are set to $1\%$ and $0.1\%$, respectively.

\begin{figure}
     \centering
     \begin{subfigure}[b]{0.49\textwidth}
         \centering
         \includegraphics[width=\textwidth]{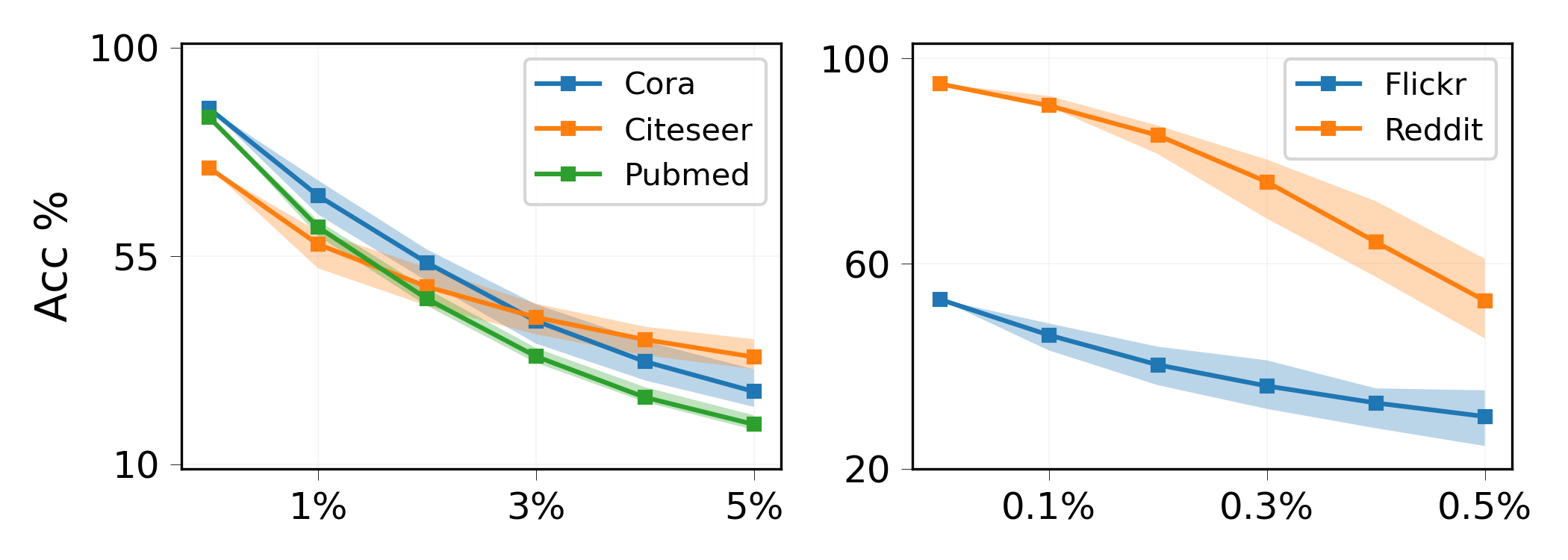}

         \caption{Budgets of node percentages}
     \end{subfigure}
     \begin{subfigure}[b]{0.49\textwidth}
         \centering
         \includegraphics[width=\textwidth]{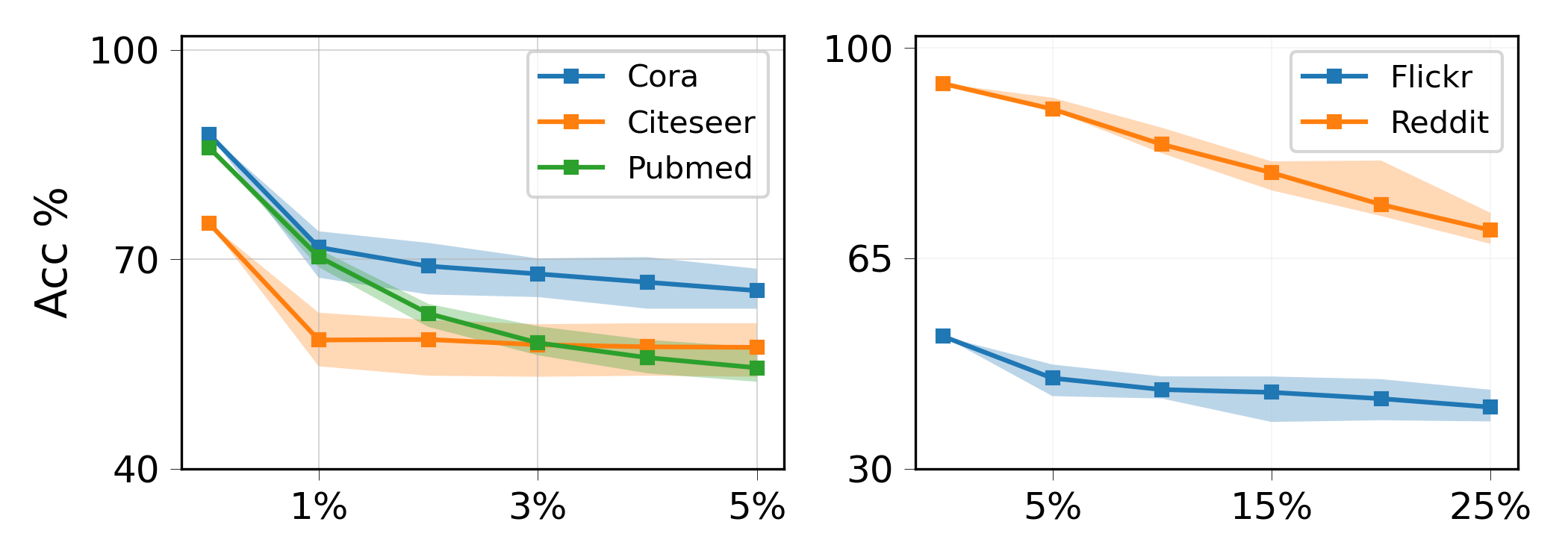}
         \caption{Budgets of feature percentages}
     \end{subfigure}
        \caption{Attacks with different perturbation budgets}
        \label{fig:params}
\end{figure}

We conduct our experiments in the generic untargeted setting and evaluate the resulting classification accuracy on the GCN model for all setups. The experimental results are summarized in Figure~\ref{fig:params}, which presents the effect of different node and feature budgets on the attack performance for each dataset.

The outcomes reveal that as the node modification rate increases, the models' classification accuracy decreases. Our attack method significantly reduces the model's performance in a rapid fashion for all datasets. 
Notably, by attacking merely $5\%$ of the graph nodes, our method can on average reduce the model's classification accuracy to $24.5\%$ for the three citation networks. Furthermore, with a scant $0.5\%$ of target nodes, we can decrease the classification accuracy to $39.2\%$ for Flickr and Reddit. Regarding feature perturbation rates, the model's performance gradually drops as the percentage increases, and stabilizes after a certain point. These outcomes suggest that only a handful of features play a pivotal role in model predictions. Simultaneously, the node perturbation rate asserts more influence than the feature perturbation rate on our attack approach's efficacy.

\begin{table*}[t!]
    \centering
    \caption{Ablation study (test accuracy \%)} 
    \label{tab:ablation}
    \resizebox{0.99\textwidth}{!}
    {
    \begin{tabular}{lccccccccc}
        \toprule
         \multicolumn{1}{l|}{\multirow{2}{*}{Method}}    & \multicolumn{3}{c|}{Cora}                                          & \multicolumn{3}{c|}{Citeseer}                                                  & \multicolumn{3}{c}{Pubmed}          \\
         \multicolumn{1}{l|}{}                           & JKNetMax          & GCN               & \multicolumn{1}{c|}{GAT}                  & JKNetMax             & GCN               & \multicolumn{1}{c|}{GAT}                & JKNetMax          & GCN               & {GAT}          \\    
        \hline
        \multicolumn{1}{l|}{$\pg{}_{global\_perturb.}$}  & 69.6 $\pm$ 3.7    & 77.7 $\pm$ 2.6    & \multicolumn{1}{c|}{66.9 $\pm$ 4.9}       & 57.4 $\pm$ 2.8       & 64.0 $\pm$ 2.2    & \multicolumn{1}{c|}{60.1 $\pm$ 3.8}     & 65.0 $\pm$ 1.0    & 72.5 $\pm$ 1.5    & 53.3 $\pm$ 0.8 \\
        \multicolumn{1}{l|}{$\pg{}_{inconsistency}$}     & 67.7 $\pm$ 2.2    & 69.9 $\pm$ 2.4    & \multicolumn{1}{c|}{60.7 $\pm$ 3.1}       & 55.9 $\pm$ 1.8       & 60.9 $\pm$ 1.4    & \multicolumn{1}{c|}{58.8 $\pm$ 1.4}     & 60.5 $\pm$ 2.2    & 65.6 $\pm$ 1.4    & 56.9 $\pm$ 1.8 \\
        \multicolumn{1}{l|}{$\pg{}_{original}$}          & 64.2 $\pm$ 2.0    & 67.4 $\pm$ 1.9    & \multicolumn{1}{c|}{61.6 $\pm$ 2.8}       & 51.3 $\pm$ 1.8       & 57.6 $\pm$ 1.9    & \multicolumn{1}{c|}{56.1 $\pm$ 1.8}     & 52.4 $\pm$ 1.0    & 61.2 $\pm$ 0.8    & 48.0 $\pm$ 1.6 \\
        \bottomrule
\end{tabular} }
\end{table*}

\subsection{Analysis of Computational Complexity}
In our method, the influence computation of one node is an LP problem. If its time complexity is denoted as $O(lp)$, the overall complexity of our algorithm will be $O(MBK\cdot lp)$, where $M$ is the number of nodes, $K$ is the number of labels, and $B$ is the averaged number of one's neighboring nodes. The inclusion of $K$ is because of the perturbation consistency in Equation~\ref{equ:adversarial_influence}. It is noteworthy that our LP problem only includes decision variables with independent range constraints. This results in a significantly reduced complexity of $O(lp)$. The solving process is shown in Section~\ref{sssec:solve_lp}.

The baseline methods RWCS, GC-RWCS, InfMax-Unif and InfMax-Norm are mainly based on Random Walk method. Their complexity is $O(M^L)$ which means the number of elementary matrix-multiplication operation. Here, $M$ is the number of nodes and $L$ stands for $L$-step random walk. It is challenging to intuitively compare our time complexity with theirs. In practice, the complexity of these methods causes scalability issues when applied to large datasets like Flickr and Reddit~\ref{tab:large_table}. This implicitly indicates that our method has a lower complexity compared to these related works. Therefore, our method is more scalable. 

\subsection{Ablation Study}

In this section, we undertake an ablation study to investigate the individual contributions of our different key components to the overall attack performance. This  helps validate the rationale behind our design. 
We develop two distinct reassembly mechanisms to further substantiate the importance of two key parts of our method. 
To do this, we arrange three types of workflows: (1) $\pg{}_{global\_perturb.}$: Instead of our customized perturbation for each node, we establish a global feature perturbation for all target nodes, analogous to the baselines. (2) $\pg{}_{inconsistency}$: In this version, we do not insist on adhering to the \textbf{perturbation consistency} when evaluating the adversarial influence score of a candidate node. Contrarily, the misclassified labels of neighboring nodes, represented by $\psi(X+\epsilon_i, A)$, can be diverse and random. (3) $\pg{}_{original}$: This represents the complete design of our method. We employ the three citation graph as the representative examples for evaluating these different settings. All three models JKNetMax, GCN and GAT are considered.
The results are in Table \ref{tab:ablation}.

We can see that these two pivotal components within our attack play crucial roles. Employing a global perturbation instead of our customized perturbation exposes weakness in the attack, resulting in a drop of success attack rate by $7.41\%$. This discrepancy stands as a primary factor contributing to our method's performance over other state-of-the-art works. Similarly, the absence of our perturbation consistency mechanism in the alternative method also reveals weakness in its attack, resulting in a success attack rate $4.12\%$ lower than our original method. This validates our claim that impulsively merging perturbations $\hat{\epsilon}_{ij}$, which cause different misclassifications of node $i$'s neighboring nodes, can compromise the effectiveness of the final perturbation on the node $i$. It highlights the importance of perturbation consistency in our attack.
Overall, we can conclude that both our perturbation customization and consistency contribute to our attack's outstanding efficacy. 

\section{Conclusion}

In this study, we propose \pg{}, a practical graph node-level attack designed for the restricted black-box setting. By strategically selecting target nodes and perturbing their features to maximize adversarial influence, our integrated approach achieves significant impact while targeting only a small number of low-degree nodes within the graph. Additionally, our method easily extends to two types of label-oriented attacks, showcasing superior performance in general untargeted and label-oriented attack scenarios. The adaptability and effectiveness of \pg{} highlight its potential to disrupt the accuracy and integrity of graph neural network models under various adversarial conditions. Its practical nature, coupled with its ability to excel in different attack scenarios, positions \pg{} as a promising tool for advancing research in graph adversarial attacks and enhancing the security of graph-based machine learning systems.  As future work, we aim to derive extensions of
our strategy to other network architectures like Transformers.

\bibliographystyle{unsrt}  
\bibliography{bibfile}

\appendix

\end{document}